%% file: root.tex
\documentclass[letterpaper, 10 pt, conference, table]{ieeeconf}  

\IEEEoverridecommandlockouts                              

\usepackage{pgfplots}
\pgfplotsset{compat=1.17}
\usepackage{amsmath}
\usepackage{booktabs}
\usepackage{overpic}
\usepackage{hyperref}
\usepackage{cleveref}
\usepackage{comment}
\usepackage{bbold}
\usepackage{balance}
\usepackage[backend=bibtex,style=ieee,minnames=1,maxnames=5,maxcitenames=1,natbib=true]{biblatex}
\usepackage{multirow,makecell}
\usepackage{subfigure}
\usepackage{etoolbox}
\usepackage{xcolor}
\usepackage{pgf}
\usepackage{hhline}
\usepackage{highlight}
\usepackage{adjustbox}
\renewcommand{\opacity}{50}

\Crefname{equation}{Eq.}{Eqs.}
\Crefname{figure}{Fig.}{Figs.}
\Crefname{table}{Tab.}{Tabs.}

\definecolor{nice-red}{HTML}{E41A1C}
\definecolor{nice-orange}{HTML}{FF7F00}
\definecolor{nice-yellow}{HTML}{FFC020}
\definecolor{nice-green}{HTML}{4DAF4A}
\definecolor{nice-blue}{HTML}{377EB8}
\definecolor{nice-purple}{HTML}{984EA3}
\definecolor{nice-grey}{HTML}{6C7A89}
\definecolor{nice-pink}{HTML}{DB5A6B}

\title{\LARGE \bf
Learning to Simulate Realistic LiDARs
}

\author{Beno\^{i}t Guillard$^{1*}$, Sai Vemprala$^{2}$, Jayesh K. Gupta$^{2}$,\\ Ondrej Miksik$^{3}$, Vibhav Vineet$^{2}$, Pascal Fua$^{1}$, Ashish Kapoor$^{2}$
\thanks{*Work done during a Microsoft Research internship}
\thanks{$^{1}$École polytechnique fédérale de Lausanne, Lausanne, Switzerland
        {\tt\small benoit.guillard@epfl.ch}}%
\thanks{$^{2}$Microsoft Research, Redmond, USA}
\thanks{$^{3}$Microsoft Mixed Reality \& AI Lab, Zurich, Switzerland}
}


\begin{document}

\maketitle
\thispagestyle{empty}
\pagestyle{empty}

\begin{abstract}
Simulating realistic sensors is a challenging part in data generation for autonomous systems, often involving carefully handcrafted sensor design, scene properties, and physics modeling. 
To alleviate this, we introduce a pipeline for data-driven simulation of a realistic LiDAR sensor. We propose a model that learns a mapping between RGB images and corresponding LiDAR features such as raydrop or per-point intensities directly from real datasets. 
We show that our model can learn to encode realistic effects such as dropped points on transparent surfaces or high intensity returns on reflective materials. 
When applied to naively raycasted point clouds provided by off-the-shelf simulator software, our model enhances the data by predicting intensities and removing points based on the scene's appearance to match a real LiDAR sensor. 
We use our technique to learn models of two distinct LiDAR sensors and use them to improve simulated LiDAR data accordingly. Through a sample task of vehicle segmentation, we show that enhancing simulated point clouds with our technique improves downstream task performance.

\end{abstract}

\input{sections/intro}

\input{sections/related}
\input{sections/method}
\input{sections/results}
\input{sections/conclusion}


\balance
\clearpage
\printbibliography


\end{document}

%% file: sections/intro.tex
\section{Introduction}

Simulation plays a major role today in the development of safe and robust autonomous systems. Compared to real world data collection and testing, which can be expensive and time consuming, simulation makes it possible to {gather} massive amounts of labeled data and environmental interactions easily. In the age of deep learning, it paves the way towards training models for autonomous systems by allowing generation of massive amounts of training data. 

Simulators have become increasingly good at rendering color imagery; e.g. rasterization-based as well as advanced physics-based rendering methods deliver realistic optical properties. While simulators are thus able to synthesize images with high visual fidelity, when it comes to non-visual classes of sensors, simulations often rely on simplified models. 
This is a major limitation because today’s autonomous systems rely on a multitude of sensors - and specifically those like LiDAR lie at the heart of a majority of them \cite{yurtsever2020survey}. 

\input{figures/teaser_fig}

Accurate LiDAR modeling is challenging due to the dependence of the output on complex properties such as material reflectance, ray incidence angle, and distance. For example, when laser rays encounter glass objects, the returns are rarely detected. 
Basic LiDAR models that exist as part of robotics simulators \cite{shah2018airsim, gazebosim} often yield simple point clouds obtained by raycasting (as seen in the bottom row, middle column of~\Cref{fig:teaser}) {which do not account for such complex yet useful properties for training LiDAR based downstream tasks and closing the sim-to-real gap}. 
Simulators such as CARLA \cite{Dosovitskiy17} do offer simulated raydrop, but this involves merely dropping points from a LiDAR point cloud at random without any modeling of material properties or physical interactions. 
It is sometimes possible to handcraft more advanced sensor models, but it involves significant human effort to study individual sensor characteristics and to accurately capture environmental interactions. 
While the idea of simulating realistic LiDAR observations from real data has been investigated by \citet{manivasagam2020lidarsim}, it requires replicating large world scenes in simulation and is hard to adapt to new sensors and scenarios or bootstrap from existing data.

\input{figures/concept}

In this paper, we introduce a data-driven pipeline for simulating realistic LiDAR sensors in an attempt to close the gap between real life and simulation. Specifically, we propose a model that can learn to drop rays where no return is expected and predict returned intensities from RGB appearance alone. To avoid the complexity of data acquisition and of high-fidelity physics-based modeling, we propose to use real data and leverage the shared statistical structure
between RGB imagery and LiDAR point clouds to learn these characteristics.

Thus, as presented in~\Cref{fig:concept}, given an RGB image as input, our model outputs 1) a binary mask that specifies where rays will be dropped and 2) predicted LiDAR intensities over the scene. We call this model a \textit{R}aydrop and \textit{I}ntensity \textit{Net}work (RINet). These outputs are then used to enhance a base point cloud, as displayed in~\Cref{fig:teaser}(bottom right). Unlike other approaches to LiDAR sensor simulation \cite{manivasagam2020lidarsim}, ours does not require any asset creation or replication of large real world scenes in simulation. Instead, we can train our model directly on existing real datasets, and predict both raydrop and intensities using a single model. 

To achieve this, we suggest a novel data representation for LiDAR as an improvement over conventional range images. Our representation densifies LiDAR points by aligning them with a corresponding RGB image space, allowing models to learn both ray drop and intensity from RGB data, whereas with range images, dropped points are already missing from the input. As raydrop can originate from several sources, we propose an objective function that is designed to isolate material-specific ray drop from random sensor noise, thus letting 
{RINet focus on learning physical properties of surfaces from the appearance}. We use our approach to enhance LiDAR point clouds obtained from the CARLA simulator~\cite{Dosovitskiy17}, and show that the resulting point clouds yield better downstream task accuracy than those obtained from today's simulators.

We summarize our key contributions below. 
\begin{enumerate}
    \item We present a LiDAR {simulation pipeline} for generating realistic LiDAR data using a model that learns LiDAR properties such as raydrop and intensities directly from RGB images. 
    \item We learn to simulate the properties of two distinct LiDAR sensors, using the Waymo dataset \cite{sun2020scalability} and the SemanticKITTI dataset \cite{behley2019iccv}. We then apply these models to LiDAR point clouds generated by the CARLA simulator \cite{Dosovitskiy17}. 
    \item We use these enhanced point clouds for a downstream task of vehicle segmentation from LiDAR data, and show that our pipeline generates data that results in higher Intersection over Union (IoU) compared to existing methods. 
\end{enumerate}

%% file: figures/teaser_fig.tex
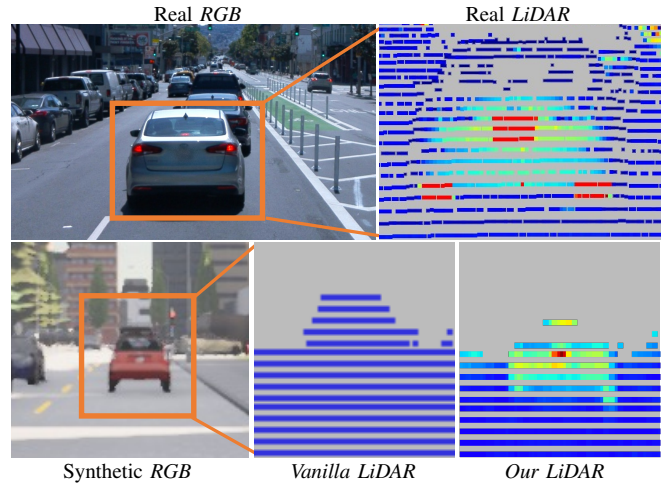
\begin{figure}
    \centering
    \vspace{2mm}
    \begin{overpic}[width=\columnwidth,trim=0 0 0 -15,clip]{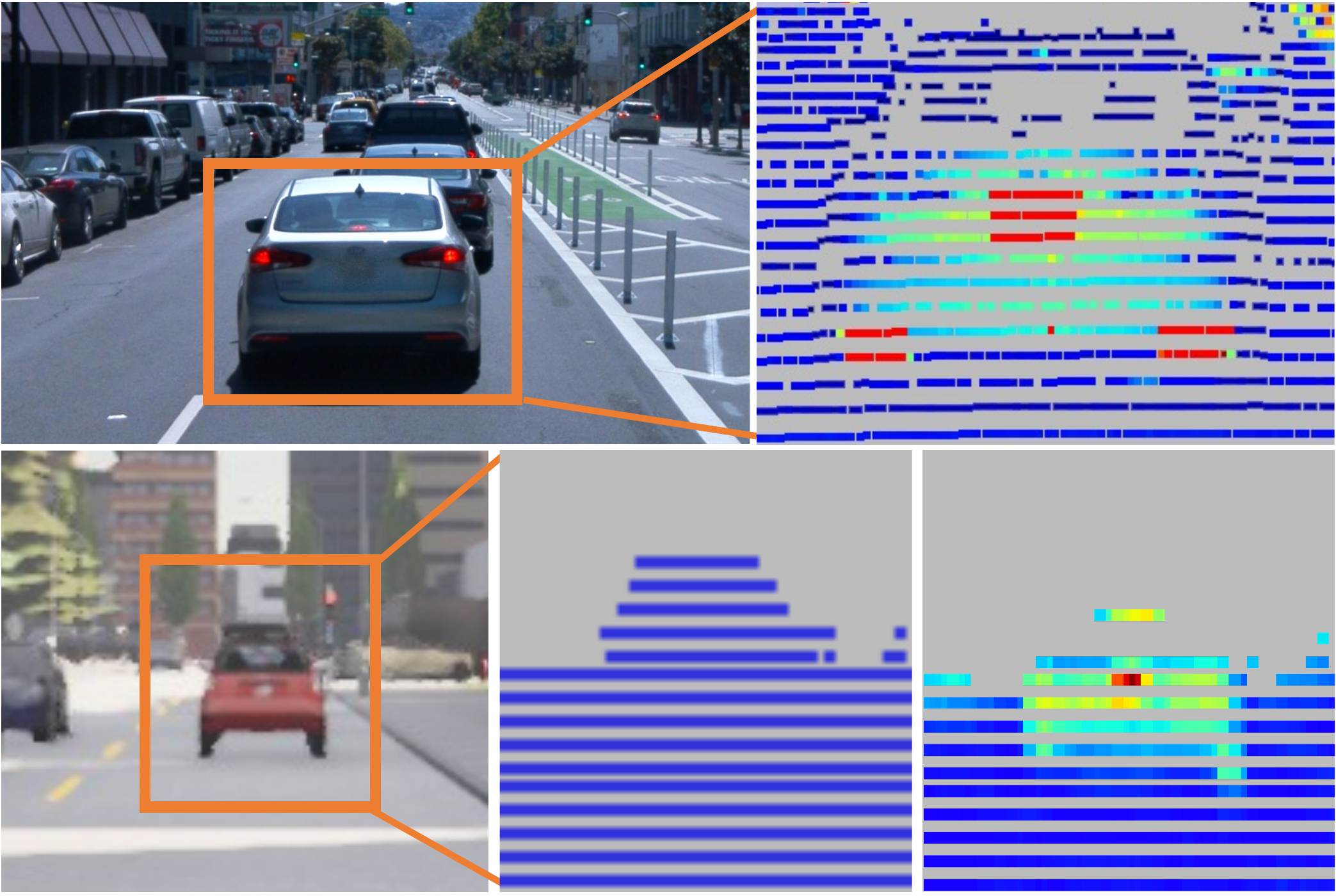}
    \put(22,67.5){\footnotesize Real \textit{RGB}}
    \put(70,67.5){\footnotesize Real \textit{LiDAR}}
    \put(8,-3){\footnotesize Synthetic \textit{RGB}}
    \put(43,-3){\footnotesize \textit{Vanilla LiDAR}}
    \put(76,-3){\footnotesize \textit{Our LiDAR}}
    \end{overpic} 
    \caption{\label{fig:teaser}\textbf{Real and synthetic LiDAR data.} \textbf{Top:} real data sample from the Waymo Perception dataset: RGB (left) and projected LiDAR colored by intensity (right). Raydrop is observed, i.e., returns are not recorded due to interactions with some surfaces such as glass. Similarly, intensity is often dependent on the type of material each ray interacts with. 
    \textbf{Bottom:} Synthetic RGB (left) and {projected} LiDAR (middle) data from CARLA's simulator. The latter does not display realistic properties.
    Through our method, we enhance basic LiDAR simulations to show  characteristics such as raydrop on glass surfaces and high intensity returns on license plates (right).}
\vspace{-5mm}    
\end{figure}

%% file: figures/concept.tex
\begin{figure*}[t]
\centering

\subfigure[Training Pipeline]{\includegraphics[width=0.4\textwidth]{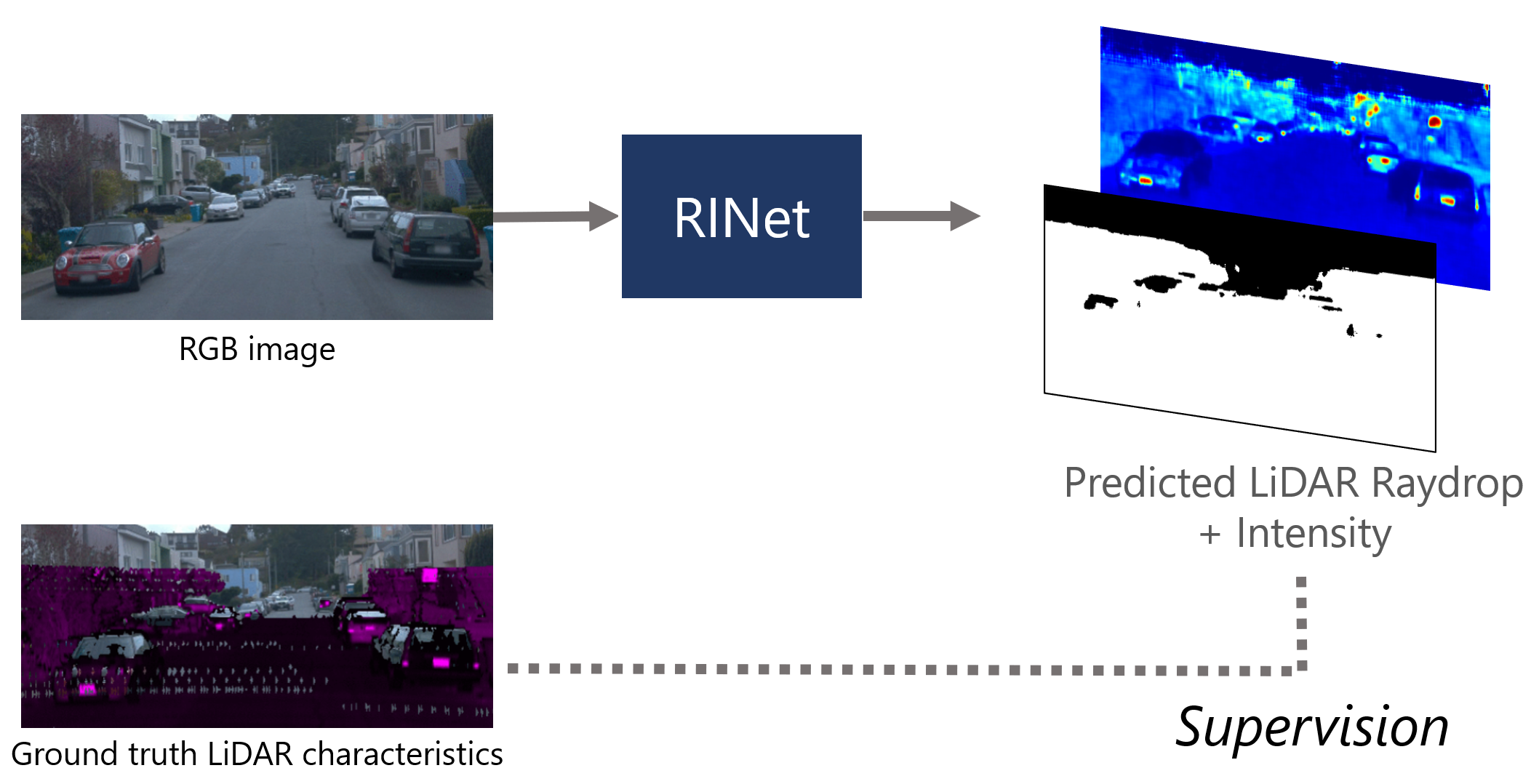}}
\subfigure[Simulation Pipeline]{\includegraphics[width=0.58\textwidth]{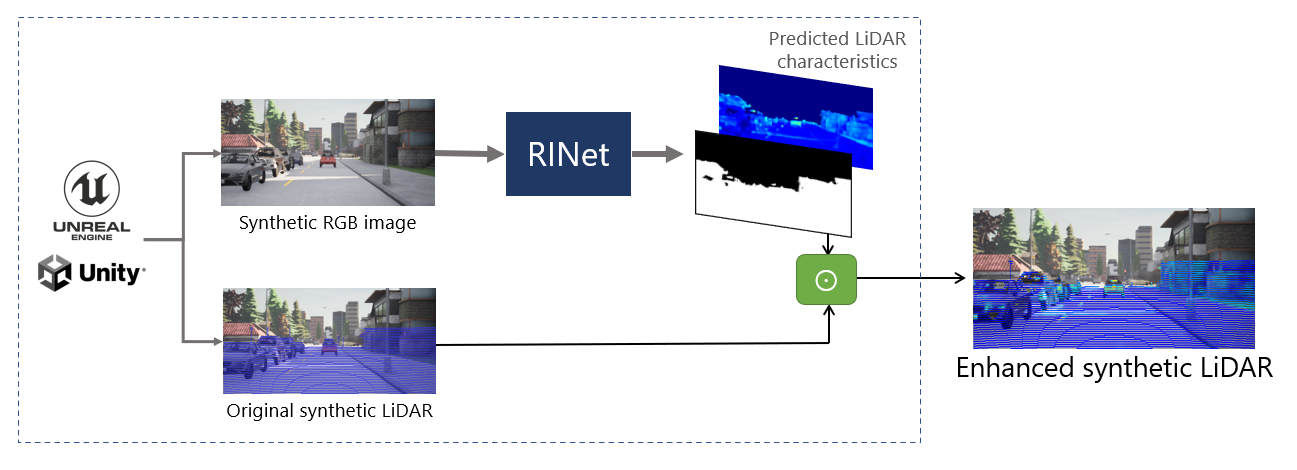}}
\caption{\label{fig:concept}\textbf{Proposed pipeline}
\textbf{(a)} In the first phase of our pipeline, we learn to predict realistic LiDAR characteristics - specifically, a binary raydrop mask and intensities directly from an RGB image, using an RINet (\textit{R}aydrop and \textit{I}ntensity \textit{Net}work).
\textbf{(b)} We use this trained RINet to enhance synthetic LiDAR data by adding intensity values and removing points based on the appearance information.}
\end{figure*}

%% file: sections/related.tex
\section{Related Work}

\subsection{LiDAR sensor principles}

The typical operating principle of a LiDAR involves scanning its field of view using one or more laser beams, and recording the returned signals through a photodetector. By processing the returned signal for each beam, LiDARs register the distance of the surface on which it was reflected, as well as an intensity value. The latter is a function of the type of material the beam interacted with, as well as the distance and the incidence angle. Furthermore, not all beams cast by the LiDAR are returned, a phenomenon called `raydrop'. Raydrop can be of two kinds: random raydrop, which is an artifact of inherent sensor noise, and physical raydrop, which is due to the beams being cast on surfaces such as glass which do not generate returns. 

Since LiDARs probe their environment with laser beams at discrete azimuth and elevation angles, a convenient representation of LiDAR data consists of using a 2D polar grid, with each row corresponding to an elevation value and each column to an azimuth. On such representations known as \textit{range images}, each pixel has 2 channels: depth (or distance), and intensity. Rays that were dropped or out of range simply yield $0$ values at the corresponding locations. An example from the Waymo Perception dataset~\cite{sun2020scalability} is shown in~\Cref{fig:data_representations}\textbf{(a)}.

\subsection{Simulating LiDAR sensors}

{Simulating LiDAR sensors is possible through several simulation software geared towards autonomous systems such as CARLA~\cite{Dosovitskiy17}, AirSim~\cite{shah2018airsim}, or Gazebo~\cite{gazebosim}. Often built over game engines such as Unity or Unreal Engine, such simulations offer basic functionality for raycasting that results in simple, dense point clouds with no intensity data. 

Prior research has also examined the idea of data-driven methods for LiDAR simulation.} LIDARSim~\cite{manivasagam2020lidarsim} involves a complex asset creation pipeline to gather digital twins of real objects that can then be composed into virtual scenes for simulations. It relies on aggregating multiple LiDAR captures and on meshing the environment to construct a digital copy of the world from which LiDAR data can be re-rendered. A neural network is trained to simulate raydrop on this digital copy of the world, but no intensity is generated. The resulting simulator only works for synthesizing LiDAR data and no other modalities; and only with objects that were captured by the original sensor and does not generalize to unseen objects.
Two other closely related approached by \citet{fang2020augmented,fang2021lidar} follow similar principles for LiDAR simulators, by placing captured or synthetic objects in virtual scenes and re-rendering them.

\citet{nakashima2021learning} and \citet{caccia2019deep} use the Generative Adversarial Network (GAN) learning framework to generate range images. The resulting networks account for raydrop but not for intensity, and provide no explicit control over the content of the newly sampled range images.

\citet{vacek2021intensities} train a Convolutional Neural Network (CNN) in a fully supervised manner on range images to learn to predict the intensity channel. As a consequence of using real range images from which rays are already missing as inputs to their network, they cannot learn raydrop. They apply their model to raycasted point clouds from a GTA game engine to produce synthetic LiDAR data with intensity. \citet{elmadawi2019end} follow a similar approach and use machine learning techniques on range images to regress Echo Pulse Width (EPW) -- a temporal equivalent to intensity. 

To close the gap between real and synthetic LiDAR data, some methods rely on deep domain adaptation using GANs~\cite{corral2022hylda} with cycle consistency~\cite{mok2021simulated, barrera2021cycle,sallab2019lidar}; or on neural style transfer~\cite{sallab2019unsupervised}. They successfully demonstrate improved downstream task performance. 

In contrast to the above methods, ours involves minimal data pre-processing, and models both raydrop and intensity with a single network, owing to a new and simple data representation. Similar to~\citet{vacek2021intensities}, we can augment already existing simulators such as~\cite{Dosovitskiy17,shah2018airsim} and {hence, one can continue to access other sensor outputs and maintain full control over} scenarios and semantic content of the scenes. As opposed to domain adaptation methods which need to be retrained for each source domain-target domain pairs, our goal is to learn a sensor network which is agnostic to the synthetic data being used. We concentrate on predicting realistic raydrop and intensity, and rely on traditional simulators to provide depth via standard raycasting.

\subsection{Learning from LiDAR data}

LiDAR sensors are used as a data source for deep-learning based methods for semantic segmentation~\cite{milioto2019iros, zhang2020polarnet}, object detection~\cite{chen2017multi, meyer2019sensor, yang20203dssd} or SLAM~\cite{chen2019suma++, chen2020rss, sun2021rsn}. These methods are trained on {annotated} real datasets such as~\cite{pandey2011ford, Geiger2012CVPR, nuscenes2019, geyer2020a2d2} and all use LiDAR intensity data except for~\citet{yang20203dssd}. Realistic LiDAR simulation allows training these models with larger amounts of generated data at minimal costs.

%% file: sections/method.tex
\section{Method}

\subsection{Task Description}
From standard autonomous system simulation software \cite{shah2018airsim, Dosovitskiy17, gazebosim} one can usually get the following data:
\begin{itemize}
    \item RGB renderings of scenes; 
    \item Clean point clouds simulating LiDAR data, but with no intensity, and no raydrop.
\end{itemize}
Our goal is to enhance such synthetic LiDAR data by:
\begin{itemize}
    \item Adding a realistic intensity channel;
    \item Removing points on surfaces that often don't return laser beams.
\end{itemize}

We propose achieving this in a data-driven fashion, 
using real world datasets which have pairs of RGB + LiDAR data and the relative poses of both sensors. Hence, given a real dataset, we train a network to predict which part of the scene would return a laser beam and the associated intensity value, based only on appearance from an RGB image. {As in CARLA,} sensor noise is simply emulated by randomly removing a fraction of points, uniformly sampled.

Thus, we define our main task as to learn the mapping from RGB appearance to LiDAR raydrop and intensity using a convolutional neural network (CNN). 
First we need to align both modalities to a common 2D spatial representation in which such a mapping can be learned, which we explain next.

\subsection{Data representation}
\input{figures/data_representations}

Previous approaches train CNNs with range images as inputs to learn LiDAR sensor characteristics such as raydrop~\cite{manivasagam2020lidarsim} or intensity~\cite{vacek2021intensities}. This representation however does not easily allow learning raydrop from appearance, because missing points are already absent from the input {data}.
For example, windows on the right car in \Cref{fig:data_representations}\textbf{(a)} do not provide any return. Learning to drop points on such surfaces (as in~\cite{manivasagam2020lidarsim}) requires to inpaint them for creating a representation in which they are not missing. This amounts to performing 3D scene reconstruction {or inpainting} -- pre-processing steps we wish to avoid. 
Moreover, the limited spatial resolution of range images (64 rows for a standard 64-channel LiDAR sensor) compared to RGB images constrains standard CNN architectures to extract coarse features only.

We therefore propose to learn typical LiDAR sensor responses in RGB camera space. Na\"{i}vely projecting LiDAR points to RGB space yields a sparse representation that is ill-suited for CNNs, as shown on \Cref{fig:data_representations}\textbf{(c)}. {Moreover, the precise locations of reprojected LiDAR points on the RGB image depend on their distance to the sensor (see~\Cref{fig:data_representations}\textbf{(c)}). A network predicting an overlay of sparse LiDAR points from RGB images is thus expected to infer the geometry of the scene. We want to avoid this, since the scene geometry can already be fully handled by the simulation software performing the raycasting operation.}

For this reason, we propose to densify LiDAR points in RGB space. 
We do so by detecting triplets of points that provided a laser return, and that are neighbors in the range image. The camera space re-projections of all such triplets are then connected with a triangle -- effectively corresponding to a meshing of LiDAR points that are neighbors on the range image. 
This yields a dense binary mask representing raydrop shown in \Cref{fig:data_representations}\textbf{(d)}, in which an empty \textit{(azimuth, elevation)} location in the range image translates into a hole. 
On this binary mask, we use bi-linear interpolation between projected points intensities to get a dense intensity channel, as shown in \Cref{fig:data_representations}\textbf{(e)}. We refer to this combination of the binary mask and the intensity channel as a \emph{dense intensity mask}.

In the following, we denote the RGB image of the scene as $I \in [0,1]^{256\times512\times3}$, and the dense intensity mask as $M \in [0,1]^{256\times512}$ where $M[u,v]=0$ means no LiDAR point was recorded for the object shown in $I$ near pixel coordinate $(u,v)$. $M$ is indeed aligned with $I$, which allows to easily learn correlations between RGB appearance and LiDAR sensor response. As detailed next, this data representation enables direct supervision of CNNs for predicting both raydrop and intensity, without involving a complex reconstruction pipeline.

\subsection{Network supervision}

We build a convolutional neural network model for predicting raydrop and intensities from RGB images, which we term RINet. 
From an RGB image $I$, RINet predicts a two channel array, where the channels correspond to two predictions: raydrop and intensities respectively.

The first channel $\widetilde{M_R} \in [0,1]^{256\times512}$ is used to supervise the network with an L1 loss on \textit{R}aydrop:
\begin{align}
\mathcal{L}_R = | \widetilde{M_R} - \mathbb{1}[M > 0] |  \; , \label{eq:loss_raydrop} 
\end{align}
where $\mathbb{1}[\cdot]$ is the element-wise indicator function.

According to~\citet{lehtinen2018noise2noise}, applying an L1 loss on noisy data allows to recover a prior on median values, and enables robust training in the presence of noise. In our case, it allows to filter out the influence of random raydrop and only predict $\widetilde{M_R}[u,v]=0$ for surfaces on which laser beams do not return to the sensor most of the time.

The second channel $\widetilde{M_I} \in [0,1]^{256\times512}$ is used to supervise the network with an L2 loss on regressed intensity values. We mask out pixels where raydrop is happening:
\begin{align}
\mathcal{L}_I = \left \| \mathbb{1}[M > 0] \odot (\widetilde{M_I} - M) \right \|_2  \; , \label{eq:loss_intensity} 
\end{align}
where $\odot$ is the element-wise multiplication.

The total loss function is taken as the sum of \Cref{eq:loss_raydrop,eq:loss_intensity}: 
$$
\mathcal{L} = \mathcal{L}_R + \mathcal{L}_I \; ,
$$
and the final predicted mask as the product of the thresholded raydrop mask and the intensity one:
\begin{equation}
\widetilde{M} =  \mathbb{1}[\widetilde{M_R} > 0.5] \odot \widetilde{M_I} \;.\label{eq:transform}
\end{equation}

\subsection{Realistic LiDAR Simulation}
\label{subsec:lidarsim}

Existing simulation engines based on Unreal Engine or Unity (e.g.\ CARLA \cite{Dosovitskiy17}, AirSim \cite{shah2018airsim} etc.) provide a rendered RGB image of the scene as well as the LiDAR point cloud obtained via simple raycasting.
RINet uses RGB images to predict realistic LiDAR phenomenon such as raydrop and intensities.
The clean raycasted pointclouds are projected to the camera space, transformation from \Cref{eq:transform} is applied, and projected back to the pointcloud space.
As desired, these points can then be assembled into range images.
Thus, \Cref{fig:concept}\textbf{(b)} describes how, with minimal invasion, we are able to leverage best parts of existing simulated data generation pipelines to enhance realism of LiDAR sensor simulations.

%% file: figures/data_representations.tex
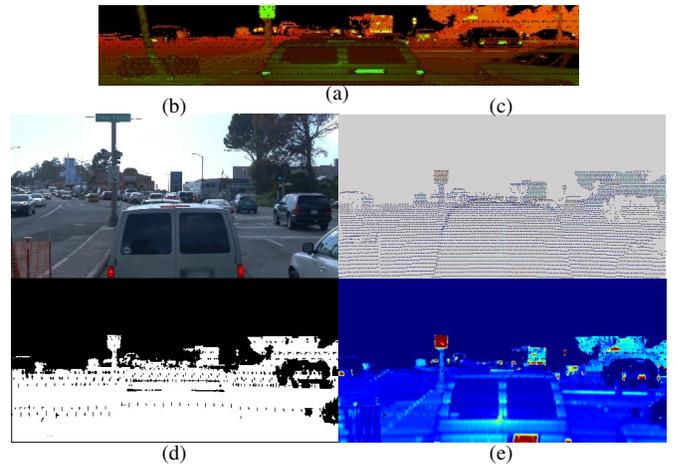
\begin{figure}[t]
\vspace{3mm}
\centering
\begin{overpic}[width=0.49\textwidth]{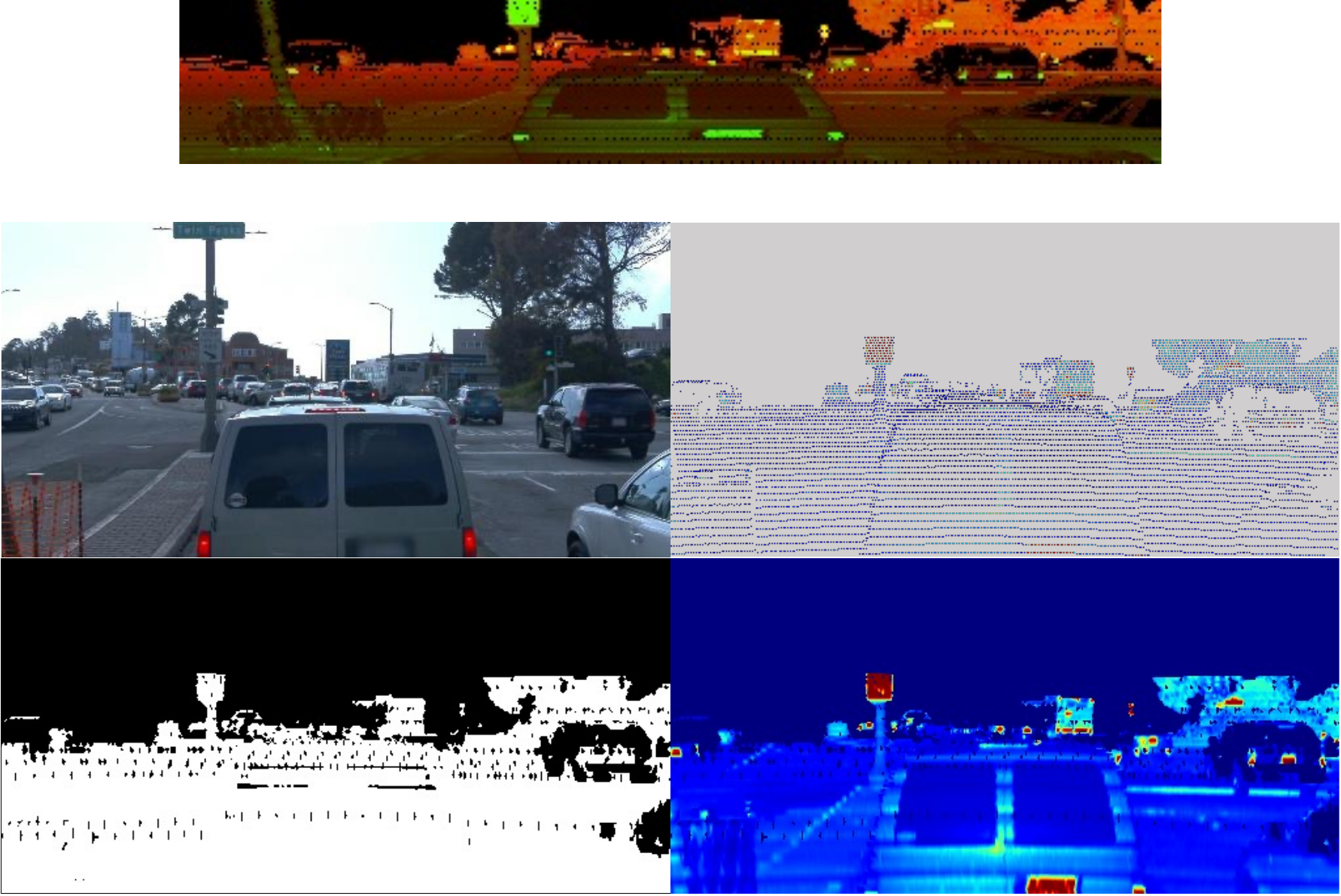}
\put(48,52.5){\footnotesize (a)}
\put(23,50.5){\footnotesize (b)}
\put(73,50.5){\footnotesize (c)}
\put(23,-2.5){\footnotesize (d)}
\put(73,-2.5){\footnotesize (e)}
 \end{overpic} \vspace{-5mm}
\caption{\label{fig:data_representations}\textbf{Data representations} 
\textbf{(a)} A LiDAR frame can be represented as a \textit{range image} (red=distance, green=intensity). Points can be projected to the RGB image of the scene \textbf{(b)}, yielding a very sparse image seen in \textbf{(c)} that is ill suited for processing by CNNs. In \textbf{(d)} we propose to densify the projection by connecting neighboring triplets of range pixels, on which we interpolate intensities to get \textbf{(e)}. It yields a mask representing intensity and raydrop which is aligned with the RGB image and more suitable for CNNs. 
\vspace{-5mm}
}
\end{figure}

%% file: sections/results.tex
\section{Results and Discussion}

In the following, we first present the implementation details of our pipeline such as datasets used and the training details. 
In Sec.~\ref{subsec:qualitative_results} we qualitatively show the ability of RINet to make relevant predictions of realistic LIDAR characteristics on synthetic data for which no ground truth is available.
Sec.~\ref{subsec:quantitative_results} quantitatively validates that synthetic data enhanced by our method can be used to train neural networks on the downstream task of LiDAR semantic segmentation. 
For training the LiDAR segmentation networks, we also experiment with adding various amounts of real labels to replicate annotation scarcity, and examine the advantages of using synthetic data in the low-data regime.  Finally in Sec.~\ref{subsec:ablations} we assess the importance of LiDAR intensities for the downstream task, the relevance of our data representation compared to range images, and discuss the effects of different levels of random noise.

\input{figures/masks_real_and_pred}

\subsection{Implementation}
We use two datasets for training the RINet model, each resulting in a distinct LiDAR model: the Waymo Perception dataset~\cite{sun2020scalability} and SemanticKITTI~\cite{behley2019iccv}. Both datasets provide sequences of paired RGB and LiDAR frames captured from a car driven in urban environments. We list some relevant details of the datasets in~\Cref{tab:dataset_details}.
Training the RINet only requires temporally synchronized RGB and LiDAR frames. In our implementation based on code from~\citet{CycleGAN2017}, RINet uses fully convolutional residual blocks~\cite{he2016deep} and instance normalization layers~\cite{ulyanov2016instance}. We use the \emph{Adam} optimizer~\cite{kingma2015adam} and train for 30 epochs, starting from a learning rate of $2e-2$ and linearly decreasing it to $0$ during the last 10 epochs.
\input{tables/dataset_characteristics}

The trained RINet models are applied on top of point clouds from CARLA using the pipeline described in \Cref{subsec:lidarsim} to provide LiDAR training data for downstream tasks. We generate 34k pairs of RGB and LiDAR frames for Waymo's sensors' configuration, and 40k for SemanticKITTI, corresponding to between 4-5k frames for 7 or 8 different CARLA environments.
We note here that CARLA allows setting parameters such as sensor pose and field of view, as desired by the user. Thus, to minimize the domain gap between real and synthetic data, when generating synthetic data corresponding to each dataset's LiDAR, we adjust the camera and LiDAR's positions and fields of view in CARLA to approximately match their real counterparts' settings.

\subsection{Simulating multiple LiDAR sensors}
\label{subsec:qualitative_results}

Given that our RINet model can be trained to simulate different LiDAR sensors, we train one sensor simulator for each real dataset: Waymo and SemanticKITTI. As shown in~\Cref{fig:masks_real_and_pred}, the simulator network picks highly reflective surfaces such as license plates, and predicts the absence of returns on transparent surfaces such as car windows. 
Notice that by design there is no sensor noise on predicted masks, thanks to the L1 supervision of~\Cref{eq:loss_raydrop}. Real masks instead display randomly distributed holes.

\input{figures/carla_clean_and_corrupted_lidar}
{In \Cref{fig:carla_clean_and_corrupted}}, we show the qualitative results of the effect of RINet's transformation of original synthetic LiDAR data based on the corresponding RGB input. Despite the domain gap introduced in input RGB images, the main features of LiDAR point clouds are preserved.
Following CARLA~\cite{Dosovitskiy17}, we {additionally} corrupt these range images by randomly masking $45$\% of the points to simulate sensor noise. 

\subsection{Relevance for downstream task}
\label{subsec:quantitative_results}
We now turn to using enhanced synthetic data generated in the manner presented above for training LiDAR segmentation networks. Labels can be obtained for free in simulated environments, thus cutting annotation costs.

We use segmentation networks on range images from~\citet{milioto2019iros} with a RangeNet21 backbone. We follow the original {training} procedure and test them on real data. We report their vehicle IoU performance on the test sets of Waymo and SemanticKITTI {which include semantic annotations of LiDAR data points}. This metric is used as a proxy to quantify the similarity between synthetic LiDAR samples and real ones, and as a direct measure of their usefulness for training purposes. Note that since we train on cropped range images and with supervision for a single object class, the metrics we report are lower than on public benchmarks.

\subsubsection{Training on synthetic data only}
\input{tables/synth_iou_comparison}
\input{figures/range_predictions}
For each one of the two sensors (Waymo and SemanticKITTI), we train a vehicle segmentation network on CARLA synthetic data only and test it on real range images. Synthetic range images are either enhanced by our method (\textit{Ours}), by simply dropping random points with a probability of $0.45$ (\textit{Noise only}), or kept as clean raycasted pointclouds (\textit{Vanilla}). \Cref{tab:synth_iou_comparison} shows the superiority of our approach compared to randomly dropping points. We train vehicle segmentation networks only with synthetic LiDAR data and test them on 2 real datasets. Enhancing synthetic training data with our pipeline (\textit{Ours}) improves performance compared to the default CARLA corruption method of random raydrop (\textit{Noise only}), which itself is better than original uncorrupted data (\textit{Vanilla}). Sample segmentation results from \textit{Vanilla} and \textit{Ours} are shown in~\Cref{fig:range_predictions}.

\subsubsection{Training on synthetic and real data}
\input{figures/synth_real_mixture}
We then turn to training with a mix of synthetic and real data, for both sensors. In addition to CARLA synthetic range images, we add $1$, $5$, $10$, $25$, $50$ or $100$\% of the real training dataset. Note that these subsets imply different ratios of synthetic vs. real data for Waymo and SemanticKITTI. \Cref{fig:synth_real_mixture} shows results for our full enhancement method (\textit{Ours}), our method without random raydrop (\textit{Ours no noise}), CARLA's default raydrop (\textit{Noise only}) and real data only (\textit{Real only}). Our method consistently outperforms CARLA's default corruption, with an average of \textbf{$\mathbf{+2.1}$\% IoU} on Waymo and \textbf{$\mathbf{+5.7}$\% IoU} on SemanticKITTI.

Our method allows for higher IoU compared to training with real data only, especially in the lower data regime. For instance, augmenting $25$\% of real training data with synthetically produced range images yields higher IoUs than training solely with $50$\% of real data. This demonstrates that synthetically produced range images can help maintain high performance levels while reducing annotation costs and labelling efforts. Performance is also improved when using $100$\% of the real training data and complementing it with synthetic range images, further proving the relevance of the enhancement process.

\Cref{fig:synth_real_mixture} also highlights that both components of the enhancement process are useful, since networks trained with our full data pipeline (\textit{Ours}) perform overall better than with each of its constituents. Indeed, solely using the data driven enhancement (\textit{Ours no noise}) or the random noise (\textit{Noise only}) separately yields lower IoUs in the majority of the cases.

\subsection{Ablation study}
\label{subsec:ablations}

\noindent \emph{1) Is intensity useful?} 
\input{tables/intensity_ablation_real}
\Cref{tab:intensity_ablation_real} shows that CNNs trained on real range images benefit from having access to the intensity channel. A segmentation network trained on Waymo range images with the intensity channel (\textit{Real}) performs better than without it (\textit{Real, no intensity}).

\input{tables/intensity_ablation_synth}
\Cref{tab:intensity_ablation_synth} shows that this is also the case for networks that are trained on synthetic range images enhanced by our pipeline. {A segmentation network trained on synthetic range images with the intensity channel (\textit{Ours}) performs better than without it (\textit{Ours, no intensity}) when tested on Waymo data. IoU degrades even more when the sensor simulator is not supervised for intensity by~\Cref{eq:loss_intensity} (\textit{Ours, no intensity, no $\mathcal{L}_I$})}. This shows that the intensity channel predicted by the LiDAR simulator is relevant and realistic. Moreover, jointly supervising the LiDAR simulator for raydrop ($\mathcal{L}_R$ in~\Cref{eq:loss_raydrop}) and intensity ($\mathcal{L}_I$ in~\Cref{eq:loss_intensity}) seems to improve the overall plausibility of the synthetic range images, further suggesting that these two aspects are entangled.

\vspace{0.5em}
\noindent \emph{2) Is our data representation better?}
\input{tables/data_representation_ablation}
To test the relevance of our upscaling of LiDAR data to camera space, we train vehicle segmentation models on two kinds of synthetic range images and test them on real Waymo data. In the first case, LiDAR data predicted by our simulator pipeline is used to supervise the downstream model. In the second case, the simulator network is a CNN trained directly on range images as in~\citet{vacek2021intensities}, without densification. To isolate the networks' contributions to generating realistic data, we do not add random raydrop noise.

\Cref{tab:data_representation_ablation} shows that our data representation allows a simulator network to provide more useful synthetic data, yielding a higher vehicle IoU of the downstream model.  We hypothesize two possible causes: the larger resolution which better preserves image features, and the ability to model physical raydrop on surfaces generating no bounces. 
{Note that the pre-processing needed to get our representation has a very light computational overhead compared to raw range images. }
However, while training RINet, since the network is processing larger arrays when using our densified masks, {from our experiments} we estimate it results in a $60$\% increase in computation time compared to range images, which are smaller.

\input{tables/noise_ablation}

\vspace{0.5em}
\noindent \emph{3) What is the ideal amount of noise?}
In the above experiments we used CARLA's default value of $p=0.45$ for the probability of randomly dropping points on synthetic range images to simulate sensor noise. In~\Cref{tab:noise_ablation} we show results of the vehicle segmentation networks trained on synthetic data from \textit{Our} pipeline and from \textit{Noise only} corruption, with different values of $p$, for both Waymo and SemanticKitti.

On both datasets, training with 1\% of real data in addition to \textit{Our} synthetic sample shows a shift in the optimal value of $p$: $p=0.1$ is optimal when not using real data for both Waymo and SemanticKITTI. With 1\% of real data, the optimal values are $0.4$ and $0.2$ respectively, suggesting a sensor or dataset-specific {behavior. }
Also note that on Waymo, \textit{Ours} always perform better that \textit{Noise only}, regardless of the value of $p$. This is only partially true on SemanticKitti, where \textit{Ours} is better for a majority of values of $p$, and fairly close to \textit{Noise only} in other cases.

Random raydrop probability is thus a hyperparameter that must be tuned depending on the application. This tuning does not involve retraining RINet, which was specifically trained to output noise-free raydrop masks.

%% file: figures/masks_real_and_pred.tex
\begin{figure}[t]
\vspace{3mm}
\centering
\includegraphics[width=0.49\columnwidth]{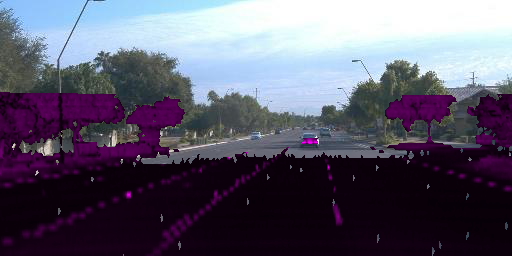} \hspace{-2mm}
\includegraphics[width=0.49\columnwidth]{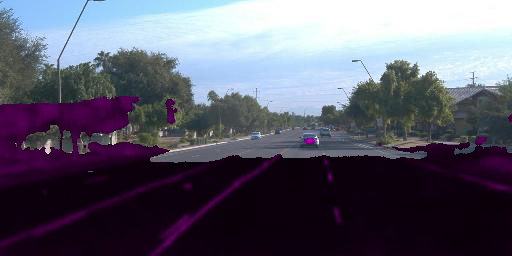}

\vspace{0.5mm}
\includegraphics[width=0.49\columnwidth]{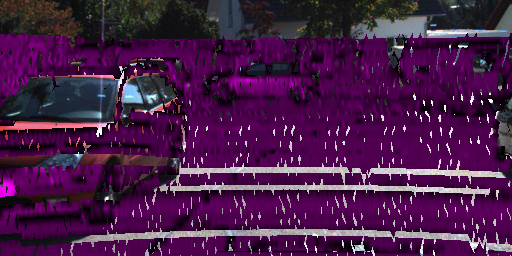} \hspace{-2mm}
\includegraphics[width=0.49\columnwidth]{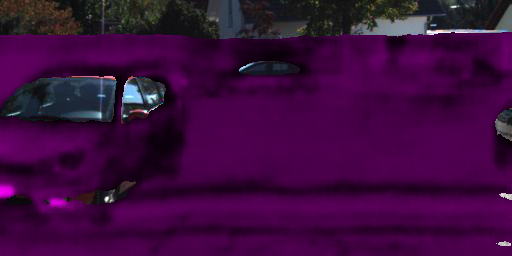}

\vspace{-3mm}
\caption{\label{fig:masks_real_and_pred}\textbf{Learning to predict intensity masks} The left column shows real intensity masks from Waymo (top) and SemanticKITTI (bottom) test sets overlaid on the RGB picture. The right column shows the corresponding predictions by RINet.}
\end{figure}

%% file: tables/dataset_characteristics.tex
\begin{table}[t]
\caption{Characteristics of datasets used for training RINet}
\label{tab:dataset_details}
\begin{adjustbox}{width=\columnwidth}
\setlength{\tabcolsep}{2pt}
\begin{tabular}{lcccccc}
\toprule
\multirow{2}*{} & LiDAR & \# of & VFoV & Max. & Training & Test \\
\multirow{2}*{} & type & channels & (deg) & range (m) & set size & set size \\ \midrule
\textbf{Waymo}         & Proprietary     & 64             & 50                 & 75             & 158081            & 39987         \\
\textbf{SemanticKITTI} & Velodyne HDL-64E & 64             & 27                 & 80             & 20409             & 2702         
\end{tabular}
\end{adjustbox}
\end{table}

%% file: figures/carla_clean_and_corrupted_lidar.tex
\begin{figure}[t]
\vspace{3mm}
\centering

\includegraphics[width=.48\textwidth]{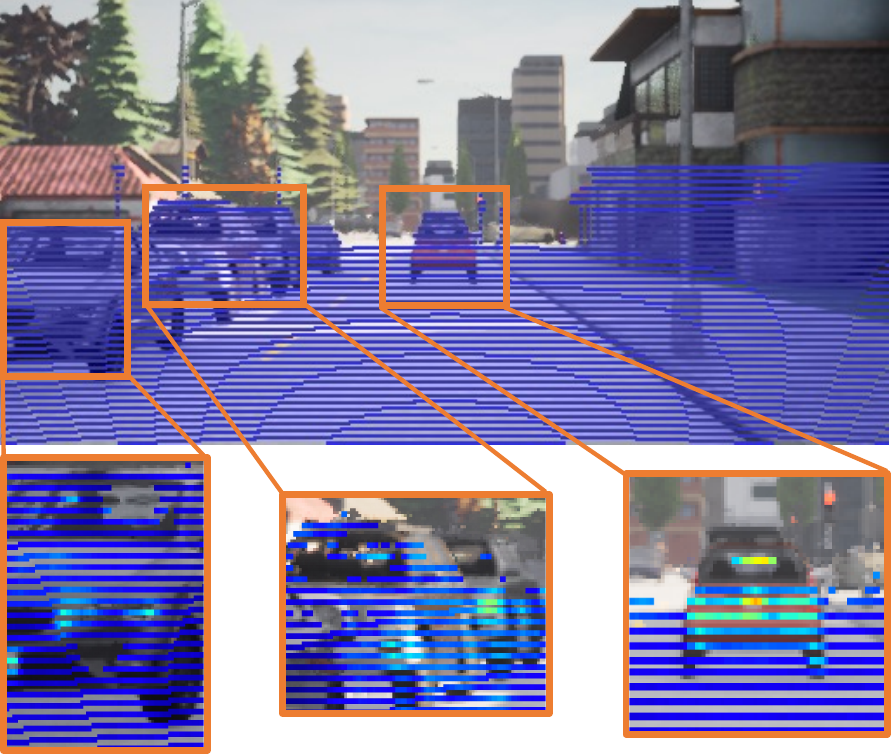}

\vspace{-2mm}
\caption{\label{fig:carla_clean_and_corrupted}\textbf{Enhancing synthetic LiDAR point clouds} CARLA pointclouds (top) have a correct geometry, but no intensity value and no raydrop. Using our method (bottom insets) we can provide the intensity information (high on car lights and plates) and remove points on surfaces that usually do not return laser beams (car windows).}
\vspace{-7mm}
\end{figure}

%% file: tables/synth_iou_comparison.tex
\begin{table}[ht]
\begin{center}
\caption{\textbf{Vehicle IoU comparison}: Training with our pipeline and baseline 
synthetic LiDARs, testing on two real datasets.}
\label{tab:synth_iou_comparison}
\begin{tabular}{rcc}
\toprule
               & Waymo  & SemanticKITTI \\ \midrule
\textit{Vanilla} & 45.7\% & 25.9\% \\
\textit{Noise only} & 53.1\% & 27.5\%        \\
\textit{Ours}  & \textbf{59.4\%} & \textbf{31.3\%}
\end{tabular}
\end{center}
\end{table}

%% file: figures/range_predictions.tex
\begin{figure*}[ht]
\vspace{3mm}
\centering

\includegraphics[width=\textwidth]{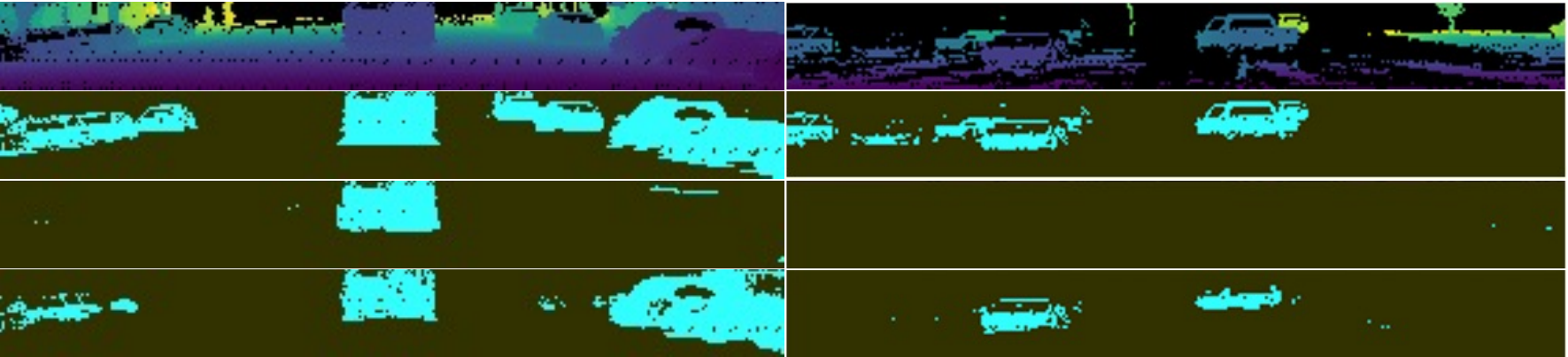}

\vspace{-3mm}
\caption{\label{fig:range_predictions}\textbf{Waymo range image segmentations} From top to bottom: \textbf{(1)} input, \textbf{(2)} ground truth vehicle segmentation masks, \textbf{(3)} predictions from a network trained with synthetic data directly from CARLA, or \textbf{(4)} from our pipeline.}
\end{figure*}

%% file: figures/synth_real_mixture.tex
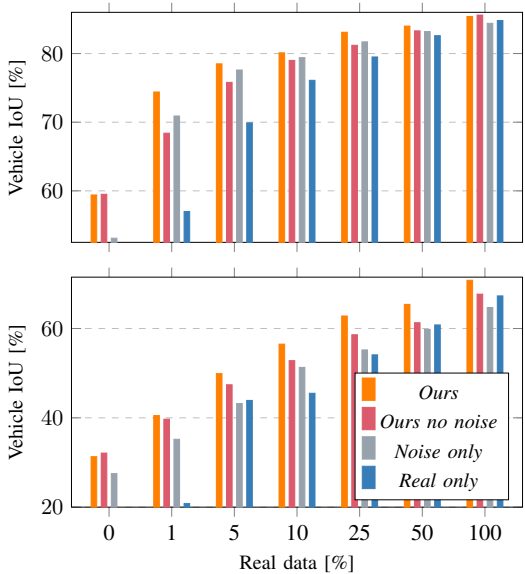
\begin{figure}[t]
\centering
\scalebox{0.85}{
\begin{tikzpicture}
\begin{axis} [
    height=0.6\columnwidth,
    width=\columnwidth,
    ybar,
    ylabel={\small Vehicle IoU [\%]},
    bar width = 2.5pt,
    xtick={0,1,2,3,4,5,6},
    xticklabels={,,},
    legend pos=south east,
    ymajorgrids=true,
    grid style=dashed,
    ymin=52.5, ymax=86,
]
 

\addplot[
    color=nice-orange,
    fill=nice-orange,
    ]
    coordinates {
    (0,59.4)(1,74.4)(2,78.5)(3,80.1)(4,83.1)(5,84.0)(6,85.4)
    };
    
\addplot[
    color=nice-pink,
    fill=nice-pink,
    ]
    coordinates {
    (0,59.5)(1,68.4)(2,75.8)(3,79)(4,81.2)(5,83.3)(6,85.6)
    };

\addplot[
    color=nice-grey!70,
    fill=nice-grey!70,
    ]
    coordinates {
    (0,53.1)(1,70.9)(2,77.6)(3,79.4)(4,81.7)(5,83.2)(6,84.4)
    };

\addplot[
    color=nice-blue,
    fill=nice-blue,
    ]
    coordinates {
    (1,57)(2,69.9)(3,76.1)(4,79.5)(5,82.6)(6,84.8)
    };
 
\end{axis}
\end{tikzpicture}
}
\scalebox{0.85}{
\begin{tikzpicture}
\begin{axis} [
    height=0.6\columnwidth,
    width=\columnwidth,
    ybar,
    xlabel={\small Real data [\%]},
    ylabel={\small Vehicle IoU [\%]},
    yticklabel pos=left,
    bar width = 2.5pt,
    xtick={0,1,2,3,4,5,6},
    xticklabels={0,1,5,10,25,50,100},
    legend pos=south east,
    ymajorgrids=true,
    grid style=dashed,
    ymin=20, ymax=71.5,
    legend image code/.code={
        \draw [#1] (0cm,-0.1cm) rectangle (0.2cm,0.25cm); }
]
 

\addplot[
    color=nice-orange,
    fill=nice-orange,
    ]
    coordinates {
    (0,31.3)(1,40.5)(2,49.9)(3,56.5)(4,62.8)(5,65.4)(6,70.8)
    };
    
\addplot[
    color=nice-pink,
    fill=nice-pink,
    ]
    coordinates {
    (0,32.1)(1,39.7)(2,47.4)(3,52.8)(4,58.6)(5,61.3)(6,67.7)
    };

\addplot[
    color=nice-grey!70,
    fill=nice-grey!70,
    ]
    coordinates {
    (0,27.5)(1,35.2)(2,43.2)(3,51.3)(4,55.2)(5,59.8)(6,64.7)
    };

\addplot[
    color=nice-blue,
    fill=nice-blue,
    ]
    coordinates {
    (1,20.8)(2,43.9)(3,45.5)(4,54.1)(5,60.8)(6,67.3)
    };
    
    \legend{\textit{\small{Ours}},
            \textit{\small{Ours no noise}},
            \textit{\small{Noise only}},
            \textit{\small{Real only}}}
 
\end{axis}
\end{tikzpicture}
}
\caption{Training on a mixture of synthetic and real data on Waymo (top) and SemanticKITTI (bottom), for different synthetic data enhancement pipelines.
}
\label{fig:synth_real_mixture}
\end{figure}

%% file: tables/intensity_ablation_real.tex
\begin{table}[t]
\begin{center}
\caption{Effect of intensity on Vehicle IoU, real data}
\label{tab:intensity_ablation_real}
\begin{tabular}{cc}
\toprule
\textit{Real}  & \textit{Real, no intensity} \\ \midrule
\textbf{84.8\%} & 83.4\%
\end{tabular}
\end{center}
\end{table}

%% file: tables/intensity_ablation_synth.tex
\begin{table}[t]
\begin{center}
\caption{Effect of Intensity on Vehicle IoU, synthetic data}
\label{tab:intensity_ablation_synth}
\begin{tabular}{ccc}
\toprule
\textit{Ours}  & \textit{Ours, no intensity} & \textit{Ours, no intensity, no $\mathcal{L}_I$} \\ \midrule
\textbf{59.4\%} & 55.9\% & 52.8\%
\end{tabular}
\end{center}
\end{table}

%% file: tables/data_representation_ablation.tex
\begin{table}[t]
\vspace{3mm}
\begin{center}
\caption{{Vehicle IoU with our representation vs. range images}}
\label{tab:data_representation_ablation}
\begin{tabular}{cc}
\toprule
\textit{Ours, no noise}  & \textit{Range image, no noise} \\ \midrule
\textbf{59.5\%} & 53.5\%

\end{tabular}
\vspace{-5mm}
\end{center}
\end{table}

%% file: tables/noise_ablation.tex
\setlength{\tabcolsep}{0pt}
\begin{table}[t]
\scriptsize
\centering
\caption{Effect of varying synthetic data noise on Vehicle IoU}
\vspace{-2mm}
\label{tab:noise_ablation}
\begin{tabular}{lrccccccc}
\Xhline{\lightrulewidth}\addlinespace[0.4em]
\multicolumn{2}{l}{\textbf{CARLA to Waymo}}          &      &      &      &      &      &      &      \\ 
\multicolumn{2}{r}{Noise probability $p=$}            & 0    & 0.1  & 0.2  & 0.3  & 0.4  & 0.45 & 0.5  \\ \hline
\multirow{2}{*}{+0\% real data} & \textit{Ours }       & \gradientcell{59.5}{56}{63}{high}{low}{50} & 
\gradientcell{62.6}{56}{63}{high}{low}{50} & 
\gradientcell{61.5}{56}{63}{high}{low}{50} & 
\gradientcell{59.9}{56}{63}{high}{low}{50} & 
\gradientcell{59.4}{56}{63}{high}{low}{50} & 
\gradientcell{59.4}{56}{63}{high}{low}{50} & 
\gradientcell{56.4}{56}{63}{high}{low}{50} \\
& \textit{Noise only } & 
\gradientcell{40.8}{40}{55}{high}{low}{50} & 
\gradientcell{50.3}{40}{55}{high}{low}{50} & 
\gradientcell{54.2}{40}{55}{high}{low}{50} & 
\gradientcell{54.1}{40}{55}{high}{low}{50} & 
\gradientcell{54.9}{40}{55}{high}{low}{50} & 
\gradientcell{53.1}{40}{55}{high}{low}{50} & 
\gradientcell{54.2}{40}{55}{high}{low}{50} \\ \hline
+1\% real data                  & \textit{Ours }       & 
\gradientcell{68.4}{68}{75.5}{high}{low}{50} & 
\gradientcell{71.9}{68}{75.5}{high}{low}{50} & 
\gradientcell{72.1}{68}{75.5}{high}{low}{50} & 
\gradientcell{73.2}{68}{75.5}{high}{low}{50} & 
\gradientcell{75.1}{68}{75.5}{high}{low}{50} & 
\gradientcell{74.4}{68}{75.5}{high}{low}{50} & 
\gradientcell{74.2}{68}{75.5}{high}{low}{50} \\
\Xhline{\lightrulewidth}\addlinespace[0.4em] 
\multicolumn{2}{l}{\textbf{CARLA to SemanticKITTI}}  &      &      &      &      &      &      &      \\
\multicolumn{2}{r}{Noise probability $p=$}            & 0    & 0.1  & 0.2  & 0.3  & 0.4  & 0.45 & 0.5  \\ \hline
\multirow{2}{*}{+0\% real data} & \textit{Ours }       & 
\gradientcell{32.1}{28.5}{34}{high}{low}{50} & 
\gradientcell{33.7}{28.5}{34}{high}{low}{50} & 
\gradientcell{33.0}{28.5}{34}{high}{low}{50} & 
\gradientcell{29.4}{28.5}{34}{high}{low}{50} & 
\gradientcell{28.8}{28.5}{34}{high}{low}{50} & 
\gradientcell{31.3}{28.5}{34}{high}{low}{50} & 
\gradientcell{31.8}{28.5}{34}{high}{low}{50} \\
& \textit{Noise only } & 
\gradientcell{25.9}{25.5}{31.5}{high}{low}{50} & 
\gradientcell{29.5}{25.5}{31.5}{high}{low}{50} & 
\gradientcell{31.1}{25.5}{31.5}{high}{low}{50} & 
\gradientcell{30.0}{25.5}{31.5}{high}{low}{50} & 
\gradientcell{29.3}{25.5}{31.5}{high}{low}{50} & 
\gradientcell{27.5}{25.5}{31.5}{high}{low}{50} & 
\gradientcell{28.6}{25.5}{31.5}{high}{low}{50} \\ \hline
+1\% real data                  & \textit{Ours }       & 
\gradientcell{39.7}{29.5}{42}{high}{low}{50} & 
\gradientcell{40.1}{29.5}{42}{high}{low}{50} & 
\gradientcell{41.8}{29.5}{42}{high}{low}{50} & 
\gradientcell{39.3}{29.5}{42}{high}{low}{50} & 
\gradientcell{40.3}{29.5}{42}{high}{low}{50} &
\gradientcell{40.5}{29.5}{42}{high}{low}{50} & 
\gradientcell{39.6}{29.5}{42}{high}{low}{50} \\
\Xhline{\lightrulewidth}
\end{tabular}
\end{table}

%% file: sections/conclusion.tex
\section{Conclusions}
We introduced a pipeline for simulating realistic LiDAR properties such as raydrop and intensities given monocular vision input. Specifically, we proposed a model RINet that learned to model LiDAR raydrop and predict intensities from RGB appearance alone. We also proposed a novel data representation that densifies LiDAR points in the RGB image space, allowing for simultaneous predictions of raydrop and intensity, while relying on basic simulators to provide depth information through simple point clouds. We evaluated our approach by modeling two different LiDAR sensors used in the Waymo and SemanticKITTI datasets to enhance LiDAR point clouds generated by the CARLA simulator. {We showed that through our pipeline, we can model LiDAR raydrop and intensities using any existing real dataset without having to re-model real world scenarios such as in~\citet{manivasagam2020lidarsim}}. Through a sample downstream task of vehicle segmentation from LiDAR data, we showed that the data 
enhanced through our pipeline result in greater segmentation IoUs compared to vanilla point clouds generated by current simulators. 

We identify several avenues for future work. Some extensions to our current method include using 360 degree imagery (when available from real datasets) to be able to simulate properties for the entire point cloud as opposed to just the forward facing region. In parallel, we plan to study the effect of scalinp up the amount of generated synthetic data. Similarly, we also wish to investigate the effect of learning from a sequence of RGB images to better model any temporal effects that could be part of the LiDAR output; {as well as more advanced ways of modeling sensor noise as opposed to a uniformly sampled one}. Currently, our method only models the RGB to LiDAR mapping, and through future work, this can be extended to a more multimodal setting where other sensor data could be leveraged in case RGB data is insufficient (for instance, night time scenarios). Furthermore, we also envisage creating a single LiDAR simulation model that can encode properties from several distinct sensors, and which can be conditioned at inference time to generate a particular kind of LiDAR data. Various neural network inference optimization techniques~\cite{cheng2017survey,liang2021pruning} could be leveraged to reduce RINet's computational expense in the simulation pipeline. Finally, since RINet is trained on real data and deployed in a simulator, its accuracy could be improved with domain adaptation training techniques or careful data augmentation.